\pdfoutput=1

\documentclass[11pt]{article}

\usepackage[]{acl}
\usepackage{booktabs}
\usepackage{times}
\usepackage{latexsym}
\usepackage{graphicx}
\usepackage{url} 
\usepackage[T1]{fontenc}

\usepackage[utf8]{inputenc}

\usepackage{microtype}

\usepackage{inconsolata}

\usepackage{multirow}
\usepackage{rotating}
\usepackage{todonotes}
\usepackage{multicol}

\usepackage[usestackEOL]{stackengine}
\usepackage{tabulary}
\usepackage[maxfloats=30,morefloats=12]{morefloats}
\usepackage{booktabs}
\usepackage{float,lscape}
\usepackage{longtable}
\usepackage{pdflscape}
\usepackage{tabularx}
\usepackage{multirow}
\usepackage{bigstrut}

\title{Are LLMs Good Cryptic Crossword Solvers?}

\author{Abdelrahman Sadallah \qquad  Daria Kotova \qquad Ekaterina Kochmar \\
Department of Natural Language Processing, MBZUAI \\
\texttt{\{abdelrahman.sadallah, daria.kotova, ekaterina.kochmar\}@mbzuai.ac.ae 
	}
}

\begin{document}
\maketitle
\begin{abstract}

Cryptic crosswords are puzzles that rely not only on general knowledge but also on the solver's ability to manipulate language on different levels and deal with various types of wordplay. Previous research suggests that solving such puzzles is a challenge even for modern NLP models. However, the abilities of large language models (LLMs) have not yet been tested on this task. In this paper, we establish the benchmark results for three popular LLMs -- {\tt LLaMA2}, {\tt Mistral}, and {\tt ChatGPT} -- showing that their performance on this task is still far from that of humans.


\end{abstract}


\section{Introduction}
\label{sec:intro}
\begin{figure}[t!]
\centering
\includegraphics[width=.45\textwidth]{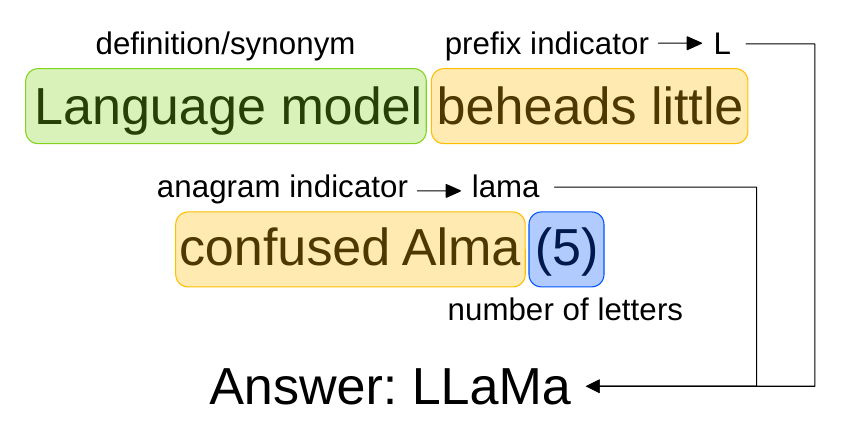}
\caption{\label{fig:example_clue} An example of a cryptic clue: number 5 at the end of the clue denotes the number of characters in the answer and is called {\bf enumeration}. The {\bf definition} part here is likely to be \textit{language model}, with the rest being the \textbf{wordplay} part. \textit{Beheads} or similar words point to the first letters of the next word, while \textit{confused} (as well as \textit{mixed up}, etc.) is likely to indicate an anagram. As we should look for a language model's name that starts with the letter \textit{l} plus an anagram of \textit{Alma} and consists of 5 letters, the answer here is \textit{Llama}.}
\label{fig:example}
\end{figure}

A cryptic crossword is a type of crossword puzzle that is known for its enigmatic clues~\citep{crosswords-definition}. Unlike standard crossword puzzles, where clues are straightforward definitions or synonyms of the answers, cryptic crosswords involve wordplay, riddles, and cleverly disguised hints that make solving them more challenging~\citep{collinsbook}. Figure \ref{fig:example} demonstrates an example of a cryptic crossword clue.


To solve a cryptic clue, one should be able to not only apply generic rules in the specific context of the clue but also use general and domain-specific knowledge to arrive at a reasonable answer. Tackling cryptic crosswords with modern NLP methods, therefore, provides an interesting challenge. It has been shown that current NLP models are far from human performance:~\citet{rozner2021decrypting}, and ~\citet{efrat2021cryptonite} report accuracy of 7.3\%, and 8.6\% for rule- and transformer-based models against 99\% achievable by expert human solvers (and 74\% by self-proclaimed amateurs) ~\citep{friedlander2009expertise, friedlander2020fluid}, and there are still no official statistics for average human performance.
This identifies a challenging area for current NLP research, while also opening up possibilities for improvement and innovation. 

\begin{table*}
\centering
\begin{tabular}{p{7cm}p{6cm}c} 
\toprule
Type        & Example Clue & Answer   \\
\cmidrule(r){1-3}
\textsc{Anagram} is a word (or words) that, when rearranged, form(s) a different word or phrase. & \underline{Never} upset \textbf{a Sci Fi writer} (5)  &  Verne \\
\textsc{Hidden clues} have the answer written in the clue itself, amongst other words. & \textbf{Confront them} \underline{in} the tob\underline{acco st}ore (6) & Accost \\
\textsc{Double definition} contains two meanings of the same word. & \textbf{In which you’d place the photo} of the \textbf{NZ author} (5)  & Frame \\
\textsc{Homophone} is a word that is pronounced the same as another but spelled differently. & \underline{Sounds like a couple} \textit{(pair)} \textbf{to scale down} (4)   & Pare \\
\bottomrule
\end{tabular}
\caption{Examples of common wordplay types. The definition part is bolded. Underlined is the part that relates to the wordplay. Actual clues do not have those indicators -- we include them here for illustrative purposes only.}
\label{tab:wordplay_types}
\end{table*}

Prior work suggests that LLMs can show emergent capabilities~\cite{wei2022emergent}, so it can be assumed that they should be able to solve cryptic puzzles if not on a par with human solvers, then at least somewhat successfully. However, to the best of our knowledge, this assumption has not been tested before. In this work, we address this research gap as we believe that trying to solve cryptic clues with LLMs might reveal their limitations as well as important aspects of natural language understanding and interpretation captured by LLMs.

Typically, a cryptic clue can be split into two parts: the \textbf{definition} and the~\textbf{wordplay}. The definition consists of one or more words in the clue that can be used interchangeably with the answer, and it usually appears either at the beginning or at the end of the clue. The wordplay can take many forms: Table \ref{tab:wordplay_types} shows the most popular wordplay types and provides an example\footnote{Examples are taken from \url{https://crypticshewrote.wordpress.com/explanations/}} for each of them. Previous work has explored explicitly splitting the solution into these two parts~\cite{rule_based_solver, rozner2021decrypting}. 

Past approaches applied to solving cryptic clues range from rule-based models,\footnote{\url{https://github.com/rdeits/cryptics}} to traditional machine learning models like KNNs \citep{rozner2021decrypting}, to Transformer models like T5 \citep{efrat2021cryptonite}. However, all these models achieve only modest accuracy on the task (\S \ref{sec:related-work}).
As language models are trained to understand language and incorporate real-world knowledge~\citep{llms-a-survey}, it can be assumed that, when prompted with a clue and short instructions, they should be able to utilize the embedded knowledge to both understand the instructions and succeed at the task. However, our preliminary investigation suggests that such a zero-shot, naive approach yields near 0\% accuracy and that LLMs struggle to understand the enumeration (answer length) constraint included in the prompt. Therefore, in this work, we experiment with prompting the models with different types of instructions  (\S \ref{ssec:prompting}), providing models with a few illustrative examples to learn from (\S \ref{ssec:few-shot}), and finally explicitly fine-tuning them on the task (\S \ref{ssec:fine-tuning}).
We use two of the most popular open-source LLMs: {\tt LLaMA2-7b}~\citep{touvron2023llama} and {\tt Mistral-7b}~\citep{jiang2023Mistral}, using QLoRA \citep{dettmers2023qlora} for Parameter Efficient Fine-Tuning (PEFT).
For comparison, we also evaluate {\tt ChatGPT}~\citep{chatgpt} -- one of the most powerful LLMs to date -- on the same task in both zero- and few-shot settings. 

Our main contributions are as follows: {\bf (1)} We explore the general abilities of LLMs on the challenging task of solving cryptic crosswords and benchmark the results of three popular LLMs; {\bf (2)} We experiment with zero- and few-shot learning as well as fine-tuning of open-source LLMs, presenting them with increasingly challenging data splits; {\bf (3)} To facilitate replicability of our results and follow-up experiments, we release our data and code.\footnote{Code and data will be released upon paper acceptance.}

\section{Related Work}
\label{sec:related-work}


\paragraph{LLMs' emergent capabilities} LLMs have been shown to follow the scaling law~\citep{kaplan2020scaling}, which has motivated researchers to explore the performance limit by increasing the size of both model and data. This has led to the discovery of the emergent abilities of LLMs~\cite{wei2022emergent}, which occur when training models with similar architectures and on the same tasks at different scales. As a result, models may exhibit unexpected abilities in solving a series of novel tasks: for instance, 
a relatively small LLM like GTP-3~\citep{gpt3} does well on arithmetic tasks, question answering or passage summarization just through in-context learning~\citep{incontext-learning}. Solving cryptic crossword puzzles is a very complex task since it requires the model to form hypotheses about the answer based on the definition and wordplay, and then select the best among such hypotheses using the information about the answer length. Whether LLMs can do this well is an open question that we aim to investigate in this work.

\paragraph{Solving puzzles with NLP models} Although there is prior work on wordplay \citep{luo2019pungan, he-etal-2019-pun, joker} and traditional crosswords \citep{LITTMAN200223, zugarini2023ratselrevolution}, much less attention has been paid to cryptic crosswords specifically. \citet{rule_based_solver} achieved 8.6\% accuracy on the task with a rule-based solver, which used the fact that a cryptic clue can be split into the definition and the wordplay parts, and applied hand-crafted probabilistic context-free grammar to find the best split. \citet{hardcastle-2007-cryptic} performed cryptic clue generation using a hand-crafted grammar.

\citet{efrat2021cryptonite} introduced {\tt Cryptonite}, a dataset of 523,114 cryptic clues collected from {\em The Times}\footnote{\url{https://www.thetimes.co.uk/puzzleclub/crosswordclub/home/crossword-cryptic}} and {\em The Telegraph}.\footnote{\url{https://puzzles.telegraph.co.uk/crossword-puzzles/cryptic-crossword}} 
They fine-tuned a T5~\citep{t5} model, which helped set the benchmark accuracy for Transformer methods at $7.6\%$. \citet{rozner2021decrypting} introduced a dataset extracted from \textit{The Guardian},\footnote{\url{https://www.theguardian.com/crosswords/series/cryptic}} and introduced a curriculum approach, which involved training a model on simpler tasks before progressing to more complex compositional clues. This increased the performance to 21.8\%.




\section{Methodology}

In order to test LLMs on their ability to solve cryptic puzzles, we challenge them in various scenarios, ranging from (1) prompting the models with various amounts of information incorporated (\S \ref{ssec:prompting}), to (2) showing the models several examples in a few-shot learning mode (\S \ref{ssec:few-shot}), to (3) explicitly fine-tuning the models on this task (\S \ref{ssec:fine-tuning}). We focus on solving the cryptic clues individually and posit that solving for the whole grid can be formulated as an iterative process of solving for each clue.

\begin{figure}[t]
    \centering  \includegraphics[width=.5\textwidth]{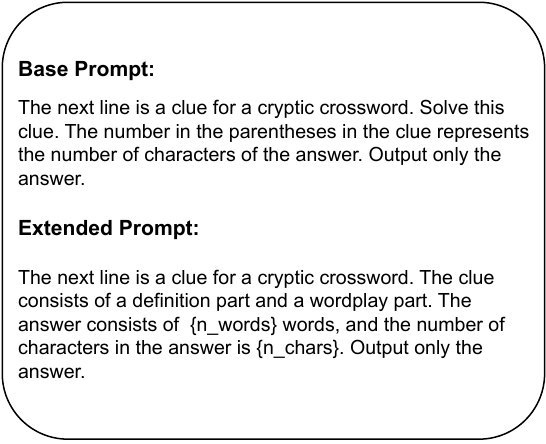}
    \caption{The two prompts that we used across all our experiments. The {\em base prompt} contains simple instruction, while the {\em extended prompt} also includes explicit information about the answer format.}
    \label{fig:prompts_variations}
\end{figure}

\subsection{Prompt variation}
\label{ssec:prompting}
As described in Section \ref{sec:intro}, prompting the models with only the crossword clue leads to 0\% accuracy, with the models not properly understanding the enumeration information given in the parentheses.
To this end, we have experimented with various prompts, incorporating additional information in the instructions for the model.
Figure \ref{fig:prompts_variations} presents two prompts that we have used in our experiments: the {\bf base prompt} is a generic definition of the task, which also includes a short hint for the model on how to interpret the enumeration part of the clue. The {\bf extended prompt}, in addition, explains the structure of the answer to the model. Experimenting with these two prompts, we investigate whether the model benefits from additional information and shows higher performance using the extended prompt.

\begin{figure}[t]
\includegraphics[width=.48\textwidth]{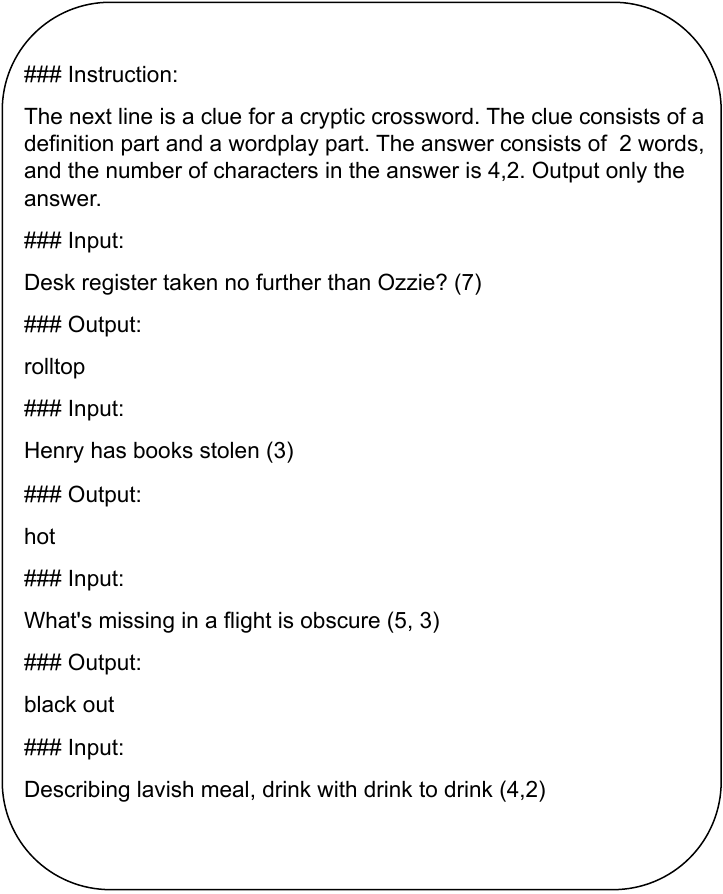}
\caption{\label{fig:few_shot} {\em 3-shot} learning input using {\em random} examples.}
\end{figure}


\subsection{Few-shot learning}
\label{ssec:few-shot}

Next, we evaluate our models using the few-shot learning technique~\cite {song2022comprehensive} using two different approaches to selecting examples for the few shots. Under the \textbf{random shots} strategy, we select examples randomly, whereas with  \textbf{indicator shots} we choose examples that are related to the current clue via an indicator word (e.g., \textit{break down} and \textit{confused}). Since the dataset does not contain information about the type of the clue or the wordplay, we use the indicator dictionary\footnote{\url{https://github.com/rdeits/cryptics/tree/master/indicators}} produced by~\citet{rule_based_solver}, which contains a list of keywords defining 6 wordplay types.
The assumption is that the models will be able to deduce from these examples, without being explicitly told, what operations they need to apply to the clue to derive the answer. This is akin to humans learning how to solve such puzzles with a few examples demonstrating the common strategies to follow. Figure \ref{fig:few_shot} shows a 3-shot example using the extended prompt.

\subsection{LLM fine-tuning}
\label{ssec:fine-tuning}
In the final setting, we explicitly fine-tune the models for our task. This step takes the models from the general language-modeling space and forces them to align with one specific task. This setting is the most computationally expensive among the three, and in order to fine-tune LLMs with limited hardware resources, we use QLoRA -- one of the recent methods of efficient fine-tuning~\cite{dettmers2023qlora} based on adapters~\cite{hu2021lora}.

\section{Data}

In our experiments, we primarily use the dataset introduced by \citet{rozner2021decrypting} and extracted from \textit{The Guardian}, as it contains data splits of different levels of difficulty and, therefore, is appropriate for testing LLMs' abilities. The dataset contains 55,783 distinct answers, resulting in an average occurrence of about 2.55 clues for each unique answer, which means that several unique but similar clues may lead to the same answer. In addition, the answers within the dataset vary in length, typically comprising 1 to 6 words, with the majority consisting of just 1 or 2 words.   

\subsection{Data splits}
\label{split}

As is highlighted in previous work~\citep{efrat2021cryptonite,rozner2021decrypting}, for this task one should select the training and test splits carefully. First of all, despite the fact that the data contains only unique clue-answer pairs, clues with the same answer still tend to be similar. Therefore, if a clue similar to the one in the training set appears in the test set with the same answer, it is likely that the model just reproduces the answer from the training, which would undermine any conclusions on the model's generalization behaviour. Secondly, \citet{rozner2021decrypting} have noted that inflections and different spelling variants of the same words can present a similar issue and suggested the \textit{word-init-disjoint} split, with which answers that share the first two characters are grouped together and allocated to a single split (training or test) only.

Although {\em word-init-disjoint} split provided by~\citet{rozner2021decrypting} takes care of the partial answer overlap, the training and test subsets may still contain several different clues with the same answer. Since this may lead to a certain level of confusion in models, we also introduce 3 new splits in addition to the two original ones proposed in~\citet{rozner2021decrypting}, which we use in our fine-tuning experiments. The splits are detailed below, with the statistics on them provided in Table \ref{tab:data_stat}:\vspace{-0.5em}
\begin{itemize}
    \item {\bf naive-random}: A shuffled random split from~\citet{rozner2021decrypting} \vspace{-0.5em}
    \item {\bf naive-random-unique}: {\em Naive-random} with only one clue for every answer (ours) \vspace{-0.5em}
    \item {\bf word-init-disjoint}: Answers that share the first two characters are grouped together and allocated  to one split (train or test) only as per~\citet{rozner2021decrypting} \vspace{-0.5em}
    \item {\bf word-init-disjoint-unique}: Same as {\em word-init-disjoint}, but with every answer occurring only once (ours) \vspace{-0.5em}
    \item {\bf word-init-disjoint-half}: Same as {\em word-init-disjoint}, but using only half of the clues with the same answer (ours)
\end{itemize}

\begin{table}
\centering
\begin{tabular}{crr} 
\toprule
\multicolumn{1}{c}{} & \multicolumn{2}{c}{Number of examples} \\
\cmidrule(rl){2-3} 
Split & Train & Test \\
\cmidrule(r){1-1} \cmidrule(rl){2-3}
naive-random & 85,428 & 28,476 \\
naive-random-unique & 47,844 & 8,444 \\
word-init-disjoint & 75,847 & 33,905 \\
word-init-disjoint-unique & 42,793 & 13,495 \\
word-init-disjoint-half & 69,339 & 21,707 \\
\bottomrule
\end{tabular}
\caption{\label{tab:data_stat} Statistics on \textit{The Guardian} dataset.}
\end{table}

\section{Experiments}
\label{sec:experiments}
To evaluate the performance of our models, in all reported experiments we use {\em accuracy} (Acc.), which is measured using exact string matching. This shows the \textit{percentage} of correctly solved cryptic clues from the datasets in question. 

In addition, previous work~\citep{efrat2021cryptonite,rozner2021decrypting} reported that enumeration provided models with useful information, while our preliminary experiments (see \S \ref{sec:intro}) suggest that LLMs struggle to benefit from this information if it is simply given as a number in the parentheses. Therefore, to further test LLMs' grasp of this information, we estimate {\em length error rate} (Len Err.) by calculating how often the length of the predicted answer differs from that of the correct answer.

\begin{table*}
\centering
\begin{tabular}{p{1.2cm}p{1.2cm}cccccccc}
\toprule
\multicolumn{2}{c}{} & \multicolumn{4}{c}{Base Prompt} & \multicolumn{4}{c}{Extended Prompt} \\
\cmidrule(rl){3-6} \cmidrule(rl){7-10}

\multirow{2}{*}{{\shortstack{Type of\\shots}}} &
    \multirow{2}{*}{{\shortstack{Number \\of shots}}}&
    \multicolumn{2}{c}{\tt LLaMA} &
    \multicolumn{2}{c}{\tt Mistral} &
    \multicolumn{2}{c}{\tt LLaMA} &
    \multicolumn{2}{c}{\tt Mistral}\\
    && Acc. & Len Err. & Acc. & Len Err. & Acc. & Len Err. & Acc. & Len Err. \\
    
\cmidrule(rl){1-2} \cmidrule(rl){3-4} \cmidrule(rl){5-6} \cmidrule(rl){7-8} \cmidrule(rl){9-10}

 & 0 & 0.2 & 89 & 0.8 & 87 & 0.2 & 89 & 0.8 & 85 \\           
\cmidrule(rl){1-10}
random & 3 & 2.4 & 81 & 3.8 & 75 & 2.2 & 81 & 3.8 & 74 \\
 & 10 & 3.0 & 81 & 4.4 & 75 & 2.8 & 81 & 4.2 & 75 \\
 \cmidrule{1-10}
indicator & 3 & 2.4 & 81 & 3.8 & 75 & 2.0 & 81 & 3.9 & 75 \\
& 10 & 2.8 & 80 & 4.4 & 75 & 2.7 & 81 & 4.3 & 75 \\
\bottomrule
\end{tabular}
\caption{\label{llm_few_shot_results}Evaluation results for {\tt LLaMA} and {\tt Mistral} on the naive-random test set using different few-shot learning scenarios. We report accuracy on cleaned outputs and length error rate for the original ones. }
\end{table*}

\begin{table}[t]
\resizebox{0.48\textwidth}{!}{%
\begin{tabular}{rcccc}
\hline
          & \multicolumn{2}{c}{Base Prompt} & \multicolumn{2}{c}{Extended Prompt} \\ \cline{2-5} 
          & Acc.       & Len Err.       & Acc.          & Len Err.         \\ \cline{2-5} 
0-shot & 6.6        & 36                 &  6.2             &    35                  \\
3-shot    & 9.5        & 35                 &   8.9            &   33                   \\ \hline
\end{tabular}%
}
\caption{{\tt ChatGPT} results using zero- and 3-shot learning with randomly sampled shots. The results with the {\em base prompt} on the {\em word-init-disjoint} split, while we used the {\em naive-random} split with the {\em extended prompt}.}
\label{tab:chat-gpt}
\end{table}

\begin{figure}
    \centering
    \includegraphics[width=.48\textwidth]{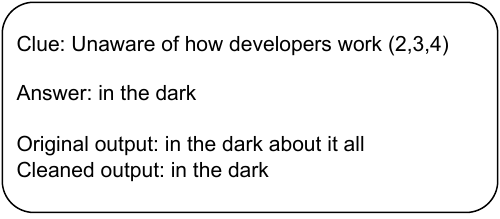}
    \caption{The full original output of the model is incorrect, however, it contains the correct answer. Using enumeration and cleaning, this answer can be extracted from the output.}
    \label{fig:cleaning_example}
\end{figure}

\subsection{Can LLMs solve the clues given various prompts?}
\label{ssec:res-prompting}

The first set of experiments, as defined in \S \ref{ssec:prompting}, considers using base and extended prompts with {\tt LLaMA} and {\tt Mistral}. The results are presented in the top row of Table \ref{llm_few_shot_results} as a zero-shot scenario since no examples are included in the prompts. Both models show similar performance with both types of prompts, suggesting that additional elaboration on the task included in the prompt itself does not help: in both cases, {\tt LLaMA} shows an accuracy of 0.2\%, while {\tt Mistral} performs marginally better and achieves an accuracy of 0.8\%. Moreover, despite the fact that both base and extended prompts include an explanation of what the numbers in parentheses mean (see Fig. \ref{fig:prompts_variations}), the models do not seem to benefit from this information, as the length error rate is above 75\% for all settings across both models. This encourages us to look into the outputs of the models further.

\paragraph{Can the models provide partially correct answers?} Enumeration provides solvers with useful information as it indicates how long the expected answer should be. Human solvers actively use this information while trying to arrive at the correct answers, and some NLP models seem to benefit from this information as well~\cite{efrat2021cryptonite,rozner2021decrypting}. However, LLMs seem to be far worse at this task: we observe a length error rate of 74-89\% across both {\tt LLaMA} and {\tt Mistral} models, with both types of prompts and under all settings (see Table \ref{llm_few_shot_results}). At the same time, a close inspection of LLMs' outputs suggests that, in fact, they often produce correct answers followed by some extraneous characters or words of various degrees of relevance (see Fig. \ref{fig:cleaning_example}). Thus, we have \textit{cleaned} the models' answers by considering only the first $n$ words for multi-word answers (where $n$ is the number of words in the correct answer), and by then only taking the first $m$ letters of the words as is specified in the enumeration part of the clue. As a result, Table \ref{llm_few_shot_results} reports the accuracy achieved by the models on cleaned outputs as well as the length error rate for the original outputs. 
Note that although cleaning helps improve the results, it does not bring the length error rate to 0\% as in some cases models produce outputs with fewer characters or words than in the correct answer.

\subsection{Few-shot learning: Do models learn from examples?}
\label{subsec:few-shots}

One way in which expert human solvers may gain valuable experience in solving cryptic puzzles is by learning from examples. In fact, it is unclear how successful human solvers are in the truly zero-shot scenario when presented with cryptic crosswords for the first time in their lives. While in \S \ref{ssec:res-prompting} we challenged our models under such zero-shot regime and showed that they do not do well, in this section we investigate whether they are capable of learning to solve the task when provided with a few (informative) examples. Figure \ref{fig:few_shot} demonstrates the use of 3 examples when prompting the model.

In Table \ref{llm_few_shot_results}, we report the results obtained with {\tt LLaMA} and {\tt Mistral} using different few-shot prompting methods. 
All these results are calculated on the {\em naive-random} test set from \citet{rozner2021decrypting} chosen because it captures a representative distribution of the data. As the results suggest, 3-shot learning with randomly selected examples ({\em random shots}) increases the model's accuracy by at least 2 percentage points (p.p.) (from 0.2 to 2.2 with the extended prompt) for {\tt LLaMA} and by 3 p.p. for {\tt Mistral}. The length error rate also decreases by at least 8 p.p. (from 89 to 81 for {\tt LLaMA}, with a larger decrease observed for {\tt Mistral}). When we increase the number of examples provided to the models from 3 to 10, we observe that the accuracy for both models increases further -- to 3.0 and 2.8 for {\tt LLaMA}, and 4.4 and 4.2 for {\tt Mistral} with the base and extended prompt, respectively. As in \S \ref{ssec:res-prompting}, we observe that (1) the extended prompt does not bring any reliable improvement, and (2) {\tt Mistral} overall outperforms {\tt LLaMA} on this task. We also conclude that both models are able to learn from the few provided examples, even when they are selected randomly, and the rate of improvement likely slows down with more provided examples.

\begin{table*}
\centering
\begin{tabular}{ccccccccc}
\toprule
 & \multicolumn{4}{c}{Base Prompt} & \multicolumn{4}{c}{Extended Prompt} \\
\cmidrule(rl){2-5} \cmidrule(rl){6-9}

& \multicolumn{2}{c}{{\tt LLaMA}} & \multicolumn{2}{c}{{\tt Mistral}} & \multicolumn{2}{c}{{\tt LLaMA}} & \multicolumn{2}{c}{{\tt Mistral}} \\

Data Split & Acc. & Len Err. & Acc. & Len Err. & Acc. & Len Err. & Acc. & Len Err.\\
\cmidrule(rl){1-1} \cmidrule(rl){2-3} \cmidrule(rl){4-5} \cmidrule(rl){6-7} \cmidrule(rl){8-9}
naive-random & 7.4 & 81 & 13.0 & 82 & 10.3 & 80 & 13.0 & 74 \\
word-init-disjoint & 0.5 & 94 & 1.2 & 80 & 0.7 & 85 & 1.2 & 89 \\
word-init-disjoint-half & 0.4 & 90 & 1.7 & 68 & 0.8 & 95 & 1.8 & 93 \\
word-init-dosjoint-unique & 0.6 & 88 & 1.7 & 91 & 0.9 & 96 & 1.5 & 80 \\
naive-random-unique & 3.8 & 92 & 8.0 & 92 & 5.0 & 96 & 8.1 & 67 \\
\bottomrule
\end{tabular}
\caption{Results obtained after fine-tuning the models with the base and extended prompts. We report accuracy on cleaned outputs and length error rate for the original ones. Data splits are described in detail in \S \ref{split}.}
\label{tab:fine-tuning_results}
\end{table*}

\paragraph{Do similar examples help?}

Next, we investigate whether more carefully selected examples, which take into account the indicator word (see \S \ref{ssec:few-shot}), help guide the models better. The results of these experiments are summarized in Table \ref{llm_few_shot_results} under {\em indicator shots}, and they do not suggest any reliable improvement: in fact, while in some cases we observe an increase of 0.1 p.p. ({\tt Mistral} with an extended prompt), in other cases model's performance even decreases ({\tt LLaMA} under several settings).

\subsection{Does ChatGPT do better?}

To get a better understanding of the difficulty of the task, we also evaluate one of the most powerful closed LLMs -- {\tt ChatGPT}. We run the evaluation for the two prompts (base and extended)\footnote{The evaluation for the extended prompt is done using only 35\% of the data, but we note that the results do not change when more data is used.} and using two test sets -- {\em naive-random} and {\em word-init-disjoint}. We observe no substantial difference in the results on these data splits and also find that providing the model with randomly selected examples in few-shot learning helps increase its performance: as is demonstrated in Table \ref{tab:chat-gpt}, accuracy improves from 6.6 to 9.5 with the base prompt. Experiments with 10-shot prompting do not suggest any significant improvements and are, therefore, not reported. These trends are aligned with those observed for the open-source models, and, although, overall, {\tt ChatGPT} shows better performance on this task, it is still significantly below human level.


\subsection{Fine-tuning: Can models be explicitly taught to solve the task?}
\label{subsec:fine-tuning}
Next, we perform fine-tuning experiments with the two open-source models -- {\tt LLaMA} and {\tt Mistral} -- using the five data splits described in \S \ref{split}. The results of these experiments are presented in Table \ref{tab:fine-tuning_results}. On their basis, we draw the following conclusions: (1) As in the previous experiments, {\tt Mistral} achieves higher accuracy than {\tt LLaMA}. (2) The overall trend that we observe is that fine-tuning and testing the models on {\em naive} splits leads to higher results across models and prompts. This suggests that the models memorize answers to a considerable extent in the {\em naive-random} split, while the performance drops rapidly by up to 5 p.p across the models on the {\em naive-random-unique} split. (3) All types of {\em disjoint} splits prove to be very challenging for both models and having different clues with similar answers apparently confuses them. Overall, we conclude that models can be taught to solve cryptic puzzles only to a limited extent.

\begin{table*}
\centering
\begin{tabular}{cccccc}
\toprule
\multicolumn{2}{c}{} & \multicolumn{2}{c}{Base Prompt} & \multicolumn{2}{c}{Extended Prompt} \\
\cmidrule(rl){3-4} \cmidrule(rl){5-6}

Train & Test & Acc. & Len Err. & Acc. & Len Err. \\
\cmidrule(rl){1-2} \cmidrule(rl){3-4} \cmidrule(rl){5-6}
cryptonite (train) & cryptonite (full set) & 14.0 & 96 & 15.7 & 99 \\
word-init-disjoint & cryptonite (full set) & 3.1 & 88 & 3.1 & 89 \\
word-init-disjoint & cryptonite (quick set) & 4.7 & 85 & 5.8 & 89 \\
\bottomrule
\end{tabular}
\caption{\label{tab:cryptonite} Results obtained with {\tt Mistral} on the {\tt Cryptonite} test set. We fine-tune {\tt Mistral} on the {\tt Cryptonite} train set. We also evaluate {\tt Mistral}, fine-tuned on our {\em word-init-disjoint} split, on the {\tt Cryptonite} full test set and quick clues subset, which  contains clues that are considered to be easier.}
\end{table*}

\subsection{Do models benefit from word masks?}



\begin{figure}
\includegraphics[width=.48\textwidth]{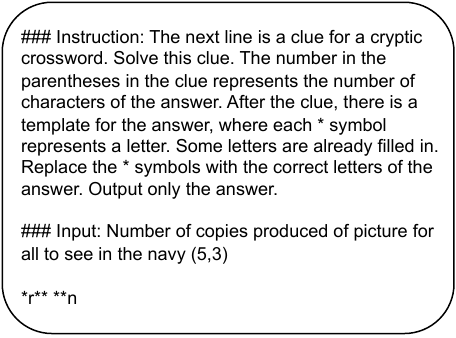}
\caption{\label{fig:spaces}Example of a prompt helping the model get the correct number of characters in the answer: the number of * symbols in the line after the clue corresponds to the number of letters in the expected answer. For further experiments, we replace some of those symbols with the correct letters of the answer in their respective positions.}
\end{figure}

Another reason why our models consistently generate answers with the wrong length might be that the prompts do not force them enough to understand the length condition. To that end, we have designed and tested an alternative prompt presented in Figure \ref{fig:spaces}, which guides the model to output words of the correct length. This is motivated by the real-world experience of solving a 2-D crossword grid, where having some already-solved clues, can provide parts of the answers for other adjacent clues.

The first line of Table \ref{tab:part_filled} reports the results of fine-tuning with the prompt without any open letters. For the next experiment, we randomly choose $p\%$ of the letters in the answers to be opened, with at least one letter opened for $p > 0$. The results for this experiment are presented in Table \ref{tab:part_filled}. We achieve the highest accuracy of 27\% so far on the {\em disjoint} split and observe that both length error rate and accuracy improve with the increase of $p$. This suggests that the model benefits from the information we release to it in our prompts.

\begin{table}
\centering
\begin{tabular}{crc} 
\toprule
Percentage filled   & Acc & Len Err.  \\
\cmidrule(r){1-3}
0 & 1.3 & 82 \\
20 & 5.1 & 70 \\
70 & 27.0 & 35 \\
\bottomrule
\end{tabular}
\caption{Results obtained by filling a certain percentage of a clue with the letters from the answer. The results are for the {\tt Mistral} model fine-tuned on the {\em word-init-disjoint} split.}
\label{tab:part_filled}
\end{table}

\begin{table}[ht]
\begin{tabular}{lcc}
\hline
Model                                  & naive-random & init-disjoint \\ \hline
\multicolumn{1}{l}{\textbf{Baselines}} &              &                    \\ \cline{1-1}

\\
Rule-based                             & 7.3          &  7.3               \\
T5 naive                               & 16.3         & 1.1                \\
T5 + curriculum                          & {\bf 21.8}         & 6.5                \\
\multicolumn{1}{l}{\textbf{Mistral}}      &              &                    \\ \cline{1-1}
Fine-tuning                    &  13.0          & 1.2                 \\
10-shot                         & 4.4          & 4.6                  \\ 
\multicolumn{1}{l}{\textbf{ChatGPT}}      &              &                    \\ \cline{1-1}
3-shot                         & 8.9          & {\bf 9.5}                  \\ \hline
\end{tabular}
\caption{Comparison between accuracy achieved by baselines, our best two models ({\tt Mistral} with the {\em extended prompt} for the fine-tuning and the {\em base prompt} for the 10-shot results), and {\tt ChatGPT} with both prompts on the test sets from ~\citet{rozner2021decrypting}. 
}
\label{tab:baselines}
\end{table}

\subsection{Experiments with Cryptonite: How generalizable are our results?}
To test the conclusions reached by~\citet{rozner2021decrypting} that the {\tt Cryptonite} dataset makes it easier for the models to memorize the answers instead of trying to solve the clues, we have fine-tuned the best open-source model from our previous experiments ({\tt Mistral}) using the data splits from~\citet{efrat2021cryptonite}. We observe that the model's performance is on a par with what we get when {\tt Mistral} is fine-tuned and tested on the {\em naive-random} split from ~\citet{rozner2021decrypting}, and this also aligns with their conclusions (see results in Table \ref{tab:cryptonite}). To further test our fine-tuned models against different subsets of the data, we fine-tune the models on the {\em word-init-disjoint} split of \textit{The Guardian} dataset and evaluate them using the full test set from {\tt Cryptonite} (after filtering the answers that overlapped with the training data used for our model), as well as the {\em quick-clues} subset, which is reported to contain simpler clues. The results in Table \ref{tab:cryptonite} suggest that  (1) {\tt Mistral} can, to a certain extent, generalize between different datasets; (2) As the accuracy obtained on {\tt Cryptonite} when the model is fine-tuned on the data from {\em word-init-disjoint} training set of \citet{rozner2021decrypting} is higher than that on the {\em word-init-disjoint} test set, {\tt Cryptonite} indeed appears to be a somewhat simpler dataset; (3) In accordance with the results from \citet{efrat2021cryptonite}, {\em quick-clues} subset is easier to solve.

\subsection{Comparison with previous baselines}
Finally, in Table \ref{tab:baselines} we compare the best results achieved by LLMs with the previous baselines introduced in~\citet{rozner2021decrypting}, including: \vspace{-0.5em}
\begin{itemize}
    \item {\bf Rule-based} -- The solver from~\citet{rule_based_solver}, which relies on the use of hand-coded grammar and WordNet~\cite{wordnet}. \vspace{-0.5em}
    \item {\bf T5 naive} -- A vanilla fine-tuned T5 model. \vspace{-0.5em}
    \item {\bf T5 + curriculum} -- A T5 model that is exposed to easier subtasks (called curricular tasks) before being fine-tuned on the main task of solving cryptic clues. 
\end{itemize}

The results suggest that open-source LLMs still do not beat the rule-based baseline on the most challenging {\em word-init-disjoint} data split, and, like Transformer-based models, suffer from the performance drop on this split. At the same time, curriculum learning improves the results of the T5 model and suggests a promising avenue for future experiments with LLMs.

\section{Conclusions and Future Work}

In this work, we have performed extensive experiments testing the abilities of two open-source ({\tt LLaMA} and {\tt Mistral}) and one closed-source ({\tt ChatGPT}) LLMs in solving challenging cryptic crossword puzzles. Our experiments encompass zero- and few-shot learning using prompts of various levels of elaboration on the task, as well as fine-tuning of the two open-source LLMs. The results suggest that, although the {\tt ChatGPT} model overall outperforms open-source LLMs, in general, cryptic crosswords still represent a very challenging task for LLMs, with a large room for improvement.



This work sets the benchmark for LLMs on the task of solving cryptic crossword puzzles, and we believe further improvements can be achieved in future work with a number of possible research directions. Firstly, a promising avenue for research in this area is chain-of-thought~\cite{wei2023chainofthought} and train-of-thought~\cite{yao2023tree} prompting techniques, which can potentially teach models how to arrive at the solution step by step.
Secondly, given a considerable increase in performance achieved by using curriculum learning with T5~\cite{rozner2021decrypting}, we consider this direction is worth exploring with LLMs as well. 
Finally, such approaches as mixture of experts~\citep{moe_1991,moe_megablock} used to train open-source models like Mixtral~\citep{jiang2024mixtral} can be applied to the task, as they may end up developing expert layers specializing in separate wordplay types.






\section*{Limitations}

\paragraph{Limited set of LLMs experimented with} Experiments with an extensive set of state-of-the-art LLMs can get quite expensive. Due to limitations of time and budget, we have been selective in terms of the LLMs that we use in this study. This has also forced us to use these LLMs in a quantized fashion. Specifically, we chose only a few of the most popular open-source and closed-source LLMs. We believe that the results obtained shed light on the current LLMs' capabilities on this task, however, we acknowledge that the set of LLMs we tested here is limited, and our results cannot be extrapolated to other LLMs. In addition, in many experiments, we have observed that certain changes in settings do not bring substantial improvement to the results (e.g., we see only relatively small improvements when we switch from 3 to 10 examples in the few-shot learning setup) -- this motivated us to perform only a limited set of experiments with some of the models in some of the settings as is elaborated in the paper.

\paragraph{Limitations of the data and the language considered} All experiments run in this study apply to data in English only. Moreover, we have experimented on only two datasets that have been released in previous publications. While this constraint is defined by the limitations of the available data, we acknowledge that our results do not necessarily extend to other languages or other datasets that may become available in the future, as this will require further experimentation.

\paragraph{Limitations of the prompting approaches} We also recognize that extensive prompt engineering may further improve the results that we report in this paper. While prompt engineering in itself was not the goal of this study and we primarily focused on the investigation of various training and test regimes for LLMs, we acknowledge that more experiments may help discover better, more informative prompts. 

\paragraph{Dangers of data contamination} Finally, we observe in our experiments that {\tt ChatGPT} outperforms the open-source models. We admit that we lack the information about its training setup, since {\tt ChatGPT} is a proprietary model, and therefore, we cannot guarantee that this model's training data is free from contamination. 


\section*{Ethics Statement}
We foresee no serious ethical implications from this study.


\bibliography{acl2020}

\appendix

\end{document}